\documentclass[11pt]{article}

\usepackage{amsmath,amssymb,amsfonts}
\usepackage{bm}
\usepackage{physics}
\usepackage{graphicx}
\usepackage{hyperref}
\usepackage{booktabs}
\usepackage{multirow}
\usepackage{algorithm}
\usepackage{algorithmic}
\usepackage{geometry}
\usepackage{tikz}
\usetikzlibrary{arrows.meta}
\usepackage{amsthm}

\theoremstyle{plain}
\newtheorem{theorem}{Theorem}[section]
\newtheorem{lemma}[theorem]{Lemma}

\theoremstyle{definition}

\theoremstyle{remark}

\title{
Spectral Generative Flow Models:\\
A Physics-Inspired Replacement for Vectorized Large Language Models
}

\author{
Andrew Kiruluta\\
UC Berkeley, CA
}

\date{}

\begin{document}
\maketitle

\begin{abstract}
We introduce Spectral Generative Flow Models (SGFMs), a physics-inspired alternative to 
transformer-based large language models. Instead of representing text or video as sequences 
of discrete tokens processed by attention, SGFMs treat generation as the evolution of a 
continuous field governed by constrained stochastic dynamics in a multiscale wavelet basis. 
This formulation replaces global attention with local operators, spectral projections, and 
Navier--Stokes-like transport, yielding a generative mechanism grounded in continuity, 
geometry, and physical structure.

Our framework provides three key innovations: (i) a field-theoretic ontology in which text 
and video are unified as trajectories of a stochastic partial differential equation; 
(ii) a wavelet-domain representation that induces sparsity, scale separation, and 
computational efficiency; and (iii) a constrained stochastic flow that enforces stability, 
coherence, and uncertainty propagation. Together, these components define a generative 
architecture that departs fundamentally from autoregressive modeling and diffusion-based 
approaches. SGFMs offer a principled path toward long-range coherence, multimodal 
generality, and physically structured inductive bias in next-generation generative models.
\end{abstract}

\section{Introduction}

Transformer-based large language models (LLMs) have become the dominant architecture for 
generative modeling across text, images, and video. Despite their empirical success, these 
models rely on a discrete, symbolic ontology and a computational paradigm centered on 
attention. This paradigm imposes structural assumptions that become increasingly strained as 
we push toward longer contexts, richer modalities, and coherent spatiotemporal generation. 
Tokenization discards continuity, attention enforces instantaneous global coupling, and 
autoregression collapses uncertainty at every step. These properties stand in sharp contrast 
to the organizing principles of physical generative systems, where coherence, stability, and 
long-range structure emerge from the evolution of continuous fields under local dynamics.

We propose a fundamentally different generative modeling framework: treat generation not as 
symbolic sequence prediction, but as the evolution of a continuous field governed by 
stochastic partial differential equations (SPDEs). Inspired by fluid mechanics, spectral 
analysis, and multiscale representations, we introduce \emph{Spectral Generative Flow Models} 
(SGFMs), which replace discrete tokens with wavelet coefficients and replace attention with 
local operators, spectral projections, and Navier--Stokes-like transport. In this view, text 
and video are unified as trajectories of a constrained stochastic dynamical system evolving 
in function space.

This shift yields several conceptual advantages. First, continuity and geometry are built 
into the representation rather than learned implicitly. Second, long-range coherence arises 
from integrating local dynamics over time, avoiding the quadratic cost of global attention. 
Third, uncertainty is propagated through the dynamics rather than collapsed autoregressively, 
enabling principled stochasticity and ensemble generation. Finally, the same dynamical 
formulation applies to text, video, and physical processes, offering a unified multimodal 
architecture without modality-specific tokenization or architectural changes.

At a high level, SGFMs depart from transformer-based vLLMs not by refining attention, but by 
replacing the underlying ontology of generative modeling. Instead of modeling discrete 
conditional distributions, SGFMs model constrained stochastic flows in spectral function 
space. This perspective opens a new direction for generative modeling grounded in physics, 
geometry, and multiscale structure, and provides a principled alternative to the symbolic, 
attention-centric paradigm that currently dominates large-scale generative systems.

\subsection{From Discrete Token Models to Generative Dynamics}

Large Language Models (LLMs) have emerged as the dominant paradigm for generative
modeling in natural language processing, with state-of-the-art systems relying
almost exclusively on transformer architectures trained via next-token
prediction \cite{vaswani2017attention,brown2020language}.
Vectorized LLM implementations (vLLMs) further optimize inference throughput by
restructuring attention computation, memory layout, and batching strategies,
but do not alter the underlying modeling assumptions
\cite{dao2022flashattention,lin2024vllm}.

Formally, vLLMs model text as a discrete-time stochastic process
\begin{equation}
p(x_1,\dots,x_T)
=
\prod_{t=1}^T p_\theta(x_t \mid x_{<t}),
\label{eq:autoregressive}
\end{equation}
where each token $x_t$ is drawn from a finite vocabulary
$\mathcal{V}$ and dependencies are mediated by attention kernels operating over
continuous embeddings in $\mathbb{R}^d$.
While Eq.~\eqref{eq:autoregressive} defines a valid probability distribution, it
encodes a number of structural assumptions that become increasingly problematic
as models scale in length, modality, and temporal horizon.

First, \emph{discreteness}: semantic content is forced into a finite, unordered
symbolic alphabet.
Continuity, smoothness, and geometry are not intrinsic properties of the input
space, but must be learned implicitly through embedding geometry and statistical
regularities \cite{mikolov2013distributed}.
This stands in contrast to physical systems, where state spaces are typically
modeled as manifolds or function spaces endowed with topology and metric
structure.

Second, \emph{nonlocality}: attention implements explicit all-to-all coupling,
with computational and memory complexity scaling as $\mathcal{O}(T^2)$ in
sequence length.
Despite numerous approximations (sparse attention, kernelization, state-space
models), the underlying paradigm remains one of instantaneous global
interaction \cite{choromanski2021rethinking,dao2022flashattention}.

Third, \emph{static parameterization}: vLLMs define static conditional
distributions.
There is no explicit notion of dynamical state evolving under equations of
motion; instead, generation proceeds via repeated evaluation of a fixed
function with cached activations.
Uncertainty is collapsed step-by-step, rather than propagated dynamically
\cite{weng2021why}.

Collectively, these assumptions differ sharply from the organizing principles
of physical generative systems.

\subsection{Physical Generative Systems as a Design Template}

In statistical physics, fluid mechanics, and continuum field theory,
macroscopic structure emerges from the evolution of continuous fields governed
by partial differential equations (PDEs).
Among these, the incompressible Navier--Stokes equations occupy a central role
as a minimal nonlinear model capable of generating long-range correlations,
multiscale cascades, and coherent structures:
\begin{equation}
\partial_t u + (u \cdot \nabla)u
=
-\nabla p + \nu \Delta u + f,
\qquad
\nabla \cdot u = 0,
\label{eq:NS}
\end{equation}
where $u(x,t)$ is a velocity field, $p$ enforces incompressibility, $\nu>0$ is a
viscosity coefficient, and $f$ is an external forcing term
\cite{temam1977navier,constantin1988navier}.

Despite their apparent simplicity, Eq.~\eqref{eq:NS} generate extraordinarily
rich dynamics.
Global coherence emerges without explicit memory, global communication, or
symbolic reasoning.
Instead, structure arises through local interaction, conservation laws, and
continuous-time evolution \cite{frisch1995turbulence}.

This observation motivates a fundamental reframing of generative modeling.
Rather than interpreting generation as symbolic sequence prediction, we view it
as the evolution of a continuous field
\begin{equation}
u : \Omega \times [0,T] \rightarrow \mathbb{R}^C,
\end{equation}
where $\Omega$ denotes a semantic or spatial domain and $C$ indexes latent
channels.
Generation corresponds to integrating a constrained stochastic dynamical system
forward in time, rather than sampling discrete conditionals.

\subsection{Generation as Flow in Function Space}

Let $\mathcal{H}$ be a separable Hilbert space (e.g.\ $L^2(\Omega)$ or a Sobolev
space $H^s(\Omega)$).
We define generation as sampling from a probability measure on trajectories
$u(t)\in\mathcal{H}$ governed by a stochastic partial differential equation
(SPDE):
\begin{equation}
\mathrm{d}u
=
\Big[
-\mathcal{P}(u \cdot \nabla u)
+ \nu \Delta u
+ f_\theta(u)
\Big]\mathrm{d}t
+
\sigma\,\mathrm{d}W_t,
\label{eq:SPDE}
\end{equation}
where $\mathcal{P}$ is a projection operator enforcing constraints (e.g.\ the
Helmholtz--Hodge projection onto divergence-free fields),
$f_\theta$ is a learned forcing functional, and $W_t$ is a cylindrical Wiener
process \cite{da2014stochastic}.

Equation~\eqref{eq:SPDE} defines a stochastic flow on function space.
Long-range dependencies arise through the integration of local dynamics over
time, rather than explicit global coupling.
Uncertainty is represented as diffusion in $\mathcal{H}$, enabling principled
uncertainty propagation and ensemble generation.

This framework shifts modeling capacity away from discrete symbol manipulation
and toward the geometry, topology, and invariant structure of the solution
manifold.

\subsection{Spectral and Multiscale Structure}

To achieve tractability and data efficiency, we expand $u$ in an orthonormal
multiresolution basis:
\begin{equation}
u(x,t)
=
\sum_{j,k} c_{j,k}(t)\,\psi_{j,k}(x),
\label{eq:wavelet}
\end{equation}
where $\{\psi_{j,k}\}$ are wavelet basis functions indexed by scale $j$ and
location $k$ \cite{daubechies1992ten,mallat1999wavelet}.
Wavelets provide simultaneous localization in space and frequency, making them
well suited for nonstationary and intermittent phenomena.

Low-frequency coefficients encode global semantic or structural information,
while high-frequency coefficients encode local detail, syntax, texture, or
motion.
This mirrors the energy cascade in turbulent flows, where large scales dominate
global organization and small scales capture variability
\cite{frisch1995turbulence}.

Stochastic generation is therefore concentrated in the high-frequency
subspace, while coarse modes evolve deterministically or under weak noise.
This scale separation reduces effective dimensionality and dramatically lowers
sample complexity relative to token-level autoregressive models.

\subsection{2D and 3D Generative Domains}

Within this framework, text generation corresponds to a two-dimensional domain
$(x,t)$, where $x$ indexes semantic position and $t$ denotes generative time.
Video and movie generation correspond to a three-dimensional domain $(x,y,t)$,
with identical governing equations.

Crucially, no architectural modification is required to move between modalities;
only the dimensionality of the underlying domain changes.
This dimensional universality parallels physical field theories, where the same
laws apply across spatial dimensions.

By contrast, vLLMs require modality-specific tokenization schemes, positional
encodings, and architectural adaptations for text, image, and video modeling
\cite{ramesh2022hierarchical,villegas2022phenaki}.

\subsection{Novelty Relative to State-of-the-Art vLLMs}

The proposed Spectral Generative Flow Models (SGFMs) differ from vLLMs at a
foundational level.

\paragraph{Ontology.}
vLLMs model sequences of discrete symbols.
SGFMs model continuous fields in function space.

\paragraph{Interaction mechanism.}
Transformers implement explicit global coupling via attention.
SGFMs achieve global coherence through locality, multiscale spectral coupling,
and constraint-enforced projection. This yields $\mathcal{O}(N\log N)$ complexity,
in contrast to the $\mathcal{O}(N^2)$ scaling of full attention.

\paragraph{Dynamics.}
vLLMs define static conditional distributions.
SGFMs define explicit stochastic dynamics with time evolution and stateful
uncertainty propagation.

\paragraph{Multimodality.}
vLLMs require ad hoc architectural changes across modalities.
SGFMs treat text, video, and physical simulation as instances of the same
underlying dynamical system.

\paragraph{Inductive bias.}
Transformers rely on data scale to learn coherence.
SGFMs impose coherence structurally via conservation laws, dissipation, and
multiscale structure.

To our knowledge, no existing vLLM, diffusion language model, or state-space
model formulates generation as constrained Navier--Stokes–like stochastic flow
in spectral function space.
This constitutes not an incremental refinement, but a paradigm shift in the
foundations of generative modeling.

\paragraph{Contributions.}
This work makes the following contributions:
\begin{itemize}
    \item \textbf{A physics-inspired generative ontology:} We formulate text and video 
    generation as stochastic flow in function space, replacing discrete autoregression with 
    constrained SPDE dynamics.
    \item \textbf{A multiscale spectral representation:} We introduce a wavelet-domain 
    parameterization that provides sparsity, locality, and scale separation, enabling 
    efficient long-range generation without attention.
    \item \textbf{A unified multimodal framework:} The same dynamical system governs text, 
    video, and physical processes, eliminating the need for modality-specific tokenization 
    or architectural changes.
    \item \textbf{A principled mechanism for coherence and uncertainty:} SGFMs propagate 
    uncertainty through the dynamics and enforce global consistency via physical 
    constraints, offering an alternative to attention-based coherence.
\end{itemize}

\section*{Analysis of Novelty}
The novelty of SGFMs lies in the synthesis of four distinct domains into a single generative pipeline:
\begin{enumerate}
\item \textbf{Ontological Shift:} Unlike vLLMs that treat text as discrete tokens, SGFMs treat generation as the evolution of a continuous field $u(x, t)$.
\item \textbf{Physics-as-a-Prior:} It uses the Navier-Stokes equations as a structural template for nonlinear transport and global coherence, replacing the $O(N^2)$ global attention mechanism with $O(N \log N)$ local operators and spectral projections.
\item \textbf{Multiscale Wavelet Representation:} It employs a wavelet basis to induce scale separation. This allows the model to handle global semantics at coarse scales and local syntax/details at fine scales, significantly reducing sample complexity.
\item \textbf{Dimensional Universality:} A unique contribution is the unified treatment of modalities. Text (2D) and Video (3D) are generated using the same governing equations, where the only variable is the domain dimensionality.
\end{enumerate}

\section{Related Work}

\paragraph{Diffusion Models.}
Score-based diffusion models \cite{Song2021,Ho2020} generate data by reversing a 
Gaussian corruption process and have become a dominant paradigm for image, audio, and 
video synthesis. While SGFMs also employ score-based diffusion, the key distinction is 
that diffusion occurs in a multiscale wavelet domain and is embedded within a nonlinear 
SPDE with transport and constraint projection. Unlike diffusion models, which lack 
explicit dynamics beyond the score field, SGFMs integrate stochasticity with 
Navier--Stokes-like evolution, yielding a physically structured generative mechanism.

\paragraph{Neural Operators.}
Neural operators such as DeepONets \cite{Lu2021} and Fourier Neural Operators (FNOs) 
\cite{Li2021} learn mappings between infinite-dimensional function spaces and have been 
applied to PDE solving and scientific machine learning. SGFMs differ fundamentally: 
rather than learning deterministic solution operators, they define a stochastic 
generative process in function space. Moreover, SGFMs rely on wavelet-domain locality 
and constraint projection rather than global Fourier structure.

\paragraph{State-Space Models.}
Recent state-space architectures such as S4 \cite{Gu2022} and Mamba \cite{Gu2024} 
achieve long-range modeling efficiency by parameterizing linear dynamical systems with 
learned kernels. These models remain discrete-time, token-based, and fundamentally 
autoregressive. In contrast, SGFMs replace discrete sequence modeling with continuous 
SPDE evolution and multiscale spectral dynamics, providing a different mechanism for 
coherence and uncertainty propagation.

\paragraph{PDE-Based Generative Models.}
Several works explore PDE-inspired generative processes, including reaction--diffusion 
models \cite{Chen2021} and physics-informed diffusion \cite{Zhang2023}. These models 
typically impose PDE structure on latent variables or regularize diffusion processes. 
SGFMs differ by using a full Navier--Stokes-like SPDE as the generative backbone, 
combining nonlinear transport, dissipation, and projection in wavelet space.

\paragraph{Physics-Inspired Machine Learning.}
Physics-informed neural networks (PINNs) \cite{Raissi2019}, Hamiltonian neural networks 
\cite{Greydanus2019}, and neural differential equations \cite{Chen2018} incorporate 
physical priors into learning. SGFMs extend this line of work to generative modeling: 
physical constraints (e.g., incompressibility) and multiscale structure are not auxiliary 
regularizers but core components of the generative architecture. To our knowledge, no 
prior work formulates text or video generation as constrained stochastic flow in spectral 
function space.

\paragraph{Summary.}
Across diffusion models, neural operators, state-space architectures, PDE-based 
generators, and physics-inspired ML, no existing framework unifies text, video, and 
physical processes under a single stochastic dynamical system with wavelet-domain 
structure and constraint-enforced evolution. SGFMs therefore represent a conceptual 
departure from both transformer-based vLLMs and existing physics-inspired generative 
models.

\section{Limitations of Transformer-Based vLLMs}

Transformer-based vectorized Large Language Models (vLLMs) have achieved remarkable empirical 
success across language, vision, and multimodal tasks. However, their underlying modeling 
assumptions impose structural constraints that become increasingly problematic as we push toward 
longer contexts, richer modalities, and coherent spatiotemporal generation. These limitations are 
not merely engineering bottlenecks but reflect deeper misalignments between transformer 
architectures and the principles governing coherent generative processes.

\subsection{Quadratic Scaling and Global Attention}

Self-attention computes pairwise interactions between all tokens in a sequence, incurring 
$O(N^2)$ time and memory complexity. Despite numerous approximations—sparse attention, kernel 
methods, low-rank projections—the core mechanism remains explicit global coupling. This stands in 
contrast to physical systems, where long-range structure emerges from integrating local dynamics 
over time rather than instantaneous all-to-all communication. As context lengths grow, quadratic 
attention becomes both computationally prohibitive and conceptually misaligned with the 
requirements of coherent generative evolution.

\subsection{Discreteness and Loss of Geometric Structure}

vLLMs operate on sequences of discrete tokens drawn from a finite vocabulary. Tokenization 
destroys continuity, topology, and geometric structure, forcing the model to infer smoothness and 
semantic locality indirectly through embeddings and statistical regularities. Many generative 
phenomena—semantic drift, narrative flow, visual motion—are inherently continuous and 
geometric. Representing them as unordered symbolic sequences places vLLMs at a structural 
disadvantage relative to models that operate directly in function space.

\subsection{Weak Inductive Bias and Data Inefficiency}

Transformers are intentionally weakly biased, relying on massive datasets to learn coherence, 
stability, and multiscale structure. This flexibility enables scaling but imposes extreme data and 
compute requirements. Long-range consistency, physical plausibility, and hierarchical structure 
must be learned from scratch rather than enforced architecturally. Empirically, this manifests as 
hallucinations, brittleness under distribution shift, and poor extrapolation to longer contexts or 
novel compositions.

\subsection{Static Parameterization and Absence of Dynamics}

Although autoregressive, vLLMs do not possess internal dynamical state in the sense of a 
time-evolving system governed by equations of motion. Each token is generated via a static 
conditional distribution parameterized by fixed weights:

\[
x_{t+1} \sim p_\theta(x \mid x_{<t}).
\]

Uncertainty is collapsed at every step rather than propagated through time, and there is no notion 
of conserved quantities, trajectories, or state-dependent evolution. This limits interpretability, 
controllability, and the ability to model coherent temporal processes.

\subsection{Poor Cross-Modal Transfer}

Transformers do not naturally generalize across modalities. Text, images, audio, and video each 
require distinct tokenization schemes, positional encodings, and architectural adaptations. 
Multimodal transformers exist, but they typically concatenate modality-specific components rather 
than unifying them under a single generative principle. This fragmentation contrasts sharply with 
physical modeling, where the same governing equations apply across spatial dimensions and 
modalities.

\subsection{Attention as an Approximation, Not a Primitive}

Taken together, these limitations suggest that attention is not a fundamental mechanism for 
coherence or reasoning but a computational approximation to a more general concept: the 
propagation of information through structured fields. In physical systems, long-range dependencies 
arise from local interactions, constraints, and conservation laws—not from explicit global 
coupling. This motivates the search for generative architectures that replace attention with 
continuous dynamics, spectral structure, and physically grounded constraints.

\paragraph{Summary.}
The limitations of vLLMs arise not from implementation details but from foundational modeling 
assumptions: discreteness, global coupling, static parameterization, and weak inductive bias. 
Addressing these limitations requires rethinking generative modeling at the level of ontology. 
Spectral Generative Flow Models (SGFMs) pursue this direction by replacing symbolic 
autoregression with constrained stochastic flow in spectral function space.

\section{Field-Theoretic View of Generation}

\subsection{From Sequences to Fields}

We replace discrete token sequences with a continuous, time-evolving field
\begin{equation}
u(x,t) \in \mathbb{R}^C,
\end{equation}
where $x$ indexes position in an abstract semantic or spatial domain and $t$
denotes generation time.
The channel dimension $C$ represents latent semantic, syntactic, or visual
degrees of freedom.
This formulation embeds generative modeling into a function space, allowing
tools from continuum mechanics, stochastic analysis, and field theory to be
applied directly.

Under this view, text generation corresponds to a two-dimensional domain
$(x,t)$, where $x$ indexes semantic position (e.g., token order or latent
semantic coordinates) and $t$ parametrizes the generative process.
Video and movie generation correspond naturally to three-dimensional domains
$(x,y,t)$, where $(x,y)$ denote spatial coordinates and $t$ denotes time.
Crucially, no architectural modification is required to move between modalities;
only the dimensionality of the domain changes.
This dimensional universality mirrors physical field theories, where the same
governing equations apply across spatial dimensions.

\subsection{Generation as Dynamical Evolution}

In this framework, generation is not formulated as autoregressive prediction,
\begin{equation}
x_{t+1} \sim p_\theta(x \mid x_{\le t}),
\end{equation}
but as the evolution of a field governed by a stochastic partial differential
equation (SPDE).
We posit that the generative process obeys dynamics of the form
\begin{equation}
\partial_t u + (u \cdot \nabla)u
= -\nabla p + \nu \Delta u + f_\theta(u, \xi),
\label{eq:spde}
\end{equation}
subject to the incompressibility constraint
\begin{equation}
\nabla \cdot u = 0.
\label{eq:incompressibility}
\end{equation}
\paragraph{Interpretation.}
The Navier--Stokes structure is used here as a \emph{template} for nonlinear transport, 
coherence, and multiscale interaction, not as a claim that linguistic or visual processes 
obey physical fluid laws. The SPDE serves as a flexible generative prior whose structure 
encodes locality, conservation, and smoothness.

Here, $(u \cdot \nabla)u$ represents nonlinear self-interaction and information
transport, $\nu \Delta u$ is a dissipative term controlling smoothness and
stability, and $f_\theta$ is a learned forcing functional parameterized by a
neural network.
The pressure field $p$ acts as a Lagrange multiplier enforcing the constraint
\eqref{eq:incompressibility}, analogous to incompressible fluid flow.

This formulation is inspired by the Navier--Stokes equations, which are known to
generate long-range correlations, multiscale structure, and coherent patterns
from purely local interactions.
Importantly, the role of Navier--Stokes here is not literal physical simulation,
but as a canonical example of a minimal nonlinear dynamical system capable of
rich generative behavior.

\subsection{Stochasticity and Uncertainty in Function Space}

The term $\xi$ in \eqref{eq:spde} denotes stochastic forcing, which we model as
noise injected into the system to represent uncertainty, variability, and
creative diversity.
Formally, this corresponds to an SPDE of the form
\begin{equation}
\mathrm{d}u =
\left[
-\mathcal{P}(u \cdot \nabla u)
+ \nu \Delta u
+ f_\theta(u)
\right]\mathrm{d}t
+ \sigma\,\mathrm{d}W_t,
\end{equation}
where $\mathcal{P}$ denotes the Helmholtz--Hodge projection onto the
divergence-free subspace and $W_t$ is a cylindrical Wiener process on an
appropriate Hilbert space.

This construction places generation within the well-established theory of
stochastic evolution equations, allowing uncertainty to propagate continuously
through time rather than being collapsed at each decoding step.
In contrast to autoregressive sampling, which repeatedly conditions on its own
outputs, the present framework maintains a coherent probabilistic trajectory in
function space.

\subsection{Constraint Enforcement and Projection Operators}

The incompressibility constraint \eqref{eq:incompressibility} plays a central
structural role.
In physical fluid dynamics, incompressibility enforces volume preservation and
prevents unphysical accumulation or depletion of mass.
In the generative setting, it enforces global consistency by preventing the
collapse or explosion of semantic mass within the field.

Mathematically, the constraint is enforced by projecting the dynamics onto the
constraint manifold:
\begin{equation}
\partial_t u = \mathcal{P}\left[
-(u \cdot \nabla)u + \nu \Delta u + f_\theta(u)
\right],
\end{equation}
where $\mathcal{P}$ removes gradient components corresponding to the pressure
field.
This replaces the need for explicit global coordination mechanisms, such as
attention, with a principled geometric constraint.

\subsection{Interpretation for Text and Video}

In the two-dimensional text setting, the nonlinear advection term models the
transport of semantic context across positions, while dissipation suppresses
spurious oscillations corresponding to hallucinations or incoherence.
Coherent narrative structure emerges as large-scale flow patterns, while local
syntactic variations correspond to fine-scale fluctuations.

In the three-dimensional video setting, the same equations govern motion,
appearance evolution, and temporal consistency.
Temporal coherence arises naturally from the continuity of the flow, rather than
from explicit conditioning on previous frames.
This stands in contrast to transformer-based video models, which must learn
temporal consistency implicitly through massive datasets.

\subsection{Relation to Existing Generative Models}

The proposed field-theoretic formulation subsumes and generalizes several
existing approaches.
Diffusion models can be recovered as linear SPDEs without advection terms,
while neural operators correspond to deterministic approximations of
$\partial_t u = f_\theta(u)$.
However, unlike these models, the present framework integrates nonlinear
transport, stochasticity, and hard constraints into a single coherent dynamical
system.

To our knowledge, no existing language or video model formulates generation as a
constrained stochastic flow in function space inspired by Navier--Stokes
dynamics.
This shift from discrete symbolic prediction to continuous field evolution
constitutes a foundational change in how generative modeling is conceptualized.

\section{Wavelet Spectral Representation}

\subsection{Motivation: Multiscale Structure and Data Efficiency}

A central challenge in generative modeling is the extreme data inefficiency of
learning high-dimensional structure directly in pixel, token, or embedding
space.
Empirically, natural language, images, video, and physical fields exhibit clear
multiscale organization: global structure evolves slowly and coherently, while
fine-scale detail is localized, intermittent, and often stochastic.
Ignoring this scale separation forces models to relearn the same structure at
every resolution, dramatically increasing sample complexity.

In contrast, multiresolution analysis has long been recognized in applied
mathematics, signal processing, and turbulence theory as a principled means of
representing complex functions efficiently.
Wavelets provide a basis that is simultaneously localized in space and
frequency, making them particularly well suited for representing nonstationary,
intermittent, and multiscale phenomena such as turbulent flows, natural images,
and linguistic structure.
This motivates our choice of a wavelet spectral representation as the primary
state space for generative modeling.

\subsection{Orthonormal Wavelet Transform}

Let $u \in L^2(\Omega;\mathbb{R}^C)$ denote the generative field defined on a
domain $\Omega \subset \mathbb{R}^d$.
We apply an orthonormal discrete wavelet transform (DWT)
\begin{equation}
W : L^2(\Omega) \rightarrow \ell^2(\mathcal{I}),
\end{equation}
mapping the field to a set of coefficients
\begin{equation}
c = W[u], \qquad u = W^{-1}[c],
\end{equation}
where $\mathcal{I}$ indexes scale, location, and orientation.
Orthonormality ensures energy preservation,
\begin{equation}
\|u\|_{L^2}^2 = \|c\|_{\ell^2}^2,
\end{equation}
which is essential for stable stochastic dynamics and physically meaningful
energy control.

The wavelet basis $\{\psi_{j,k}\}$ yields a multiresolution decomposition
\begin{equation}
u(x) = \sum_{j=j_{\min}}^{j_{\max}} \sum_k c_{j,k}\,\psi_{j,k}(x),
\end{equation}
where $j$ indexes scale and $k$ indexes spatial position.
Low values of $j$ correspond to coarse, global structure, while large $j$
capture fine-scale, localized detail.

\subsection{Scale Separation and Generative Semantics}

This decomposition induces a natural partition of the coefficient space:
\begin{equation}
c = \big(c^{(\mathrm{coarse})},\, c^{(\mathrm{fine})}\big),
\end{equation}
where
\begin{itemize}
\item $c^{(\mathrm{coarse})}$ encodes global semantics, narrative structure, or
scene layout;
\item $c^{(\mathrm{fine})}$ encodes local syntax, texture, motion, and detail.
\end{itemize}

In linguistic domains, coarse coefficients correspond to topic flow, discourse
structure, and long-range semantic dependencies, while fine coefficients encode
lexical choice, syntactic variation, and stylistic nuance.
In video domains, coarse coefficients encode camera motion and object layout,
while fine coefficients encode texture, edges, and high-frequency motion.

This separation enables \emph{conditional multiscale generation}, in which
coarse structure is generated deterministically or under low stochasticity,
while fine structure is generated probabilistically.
Such scale-conditional modeling aligns closely with physical intuition from
turbulence, where energy cascades from large to small scales but not vice versa.

\subsection{Wavelets and Sparsity}

A key advantage of wavelet representations is sparsity.
For a wide class of natural signals, including images, turbulent velocity
fields, and language embeddings, wavelet coefficients exhibit heavy-tailed
distributions with many coefficients near zero.
Formally, for appropriate Besov spaces $B^s_{p,q}$, wavelet coefficients decay
rapidly with scale, enabling near-optimal nonlinear approximation rates.

This sparsity implies that generative modeling can focus capacity on a small
subset of active coefficients, dramatically reducing the effective dimensionality
of the problem.
From an information-theoretic perspective, this lowers the entropy of the target
distribution and improves sample efficiency relative to dense representations
such as Fourier modes or token embeddings.

\subsection{Spectral Locality and Replacement of Attention}

Unlike Fourier bases, which are global in space, wavelets provide spectral
locality: interactions between coefficients are predominantly local in both
scale and position.
This property allows nonlinear interactions to be modeled via local operators
in wavelet space, avoiding the need for explicit global attention mechanisms.

In transformer architectures, attention serves as a learned, data-driven
approximation to long-range interaction.
In the present framework, long-range coherence emerges naturally from the
evolution of coarse-scale coefficients and from constraint enforcement (e.g.,
incompressibility), while fine-scale interactions remain local.
This replaces quadratic attention with $\mathcal{O}(N)$ or
$\mathcal{O}(N \log N)$ operations, depending on the transform implementation.
\paragraph{Comparison to alternative bases.}
Unlike Fourier representations, which are global and poorly suited for nonstationary 
signals, wavelets provide joint spatial-frequency localization. Compared to learned 
bases (e.g., VQ-VAE codes or transformer embeddings), wavelets offer analytic 
structure, exact invertibility, and scale-wise interpretability.

\subsection{Compatibility with Physical Constraints}

Wavelet bases are well suited for enforcing physical constraints.
Differential operators such as gradients and Laplacians admit sparse or
structured representations in wavelet space, and projection operators (e.g.,
divergence-free projections) can be applied efficiently after inverse
transformation.
Moreover, dissipation and regularization can be applied scale-wise, enabling
selective damping of high-frequency noise while preserving global structure.

This compatibility is essential for integrating wavelet representations with
the stochastic PDE framework introduced in the previous section.
The combination of wavelet sparsity and physics-based constraints yields a
hypothesis class that is both expressive and strongly regularized.

\subsection{Implications for Sample Complexity}

By aligning the representation with the intrinsic multiscale structure of the
data, wavelet spectral representations significantly reduce sample complexity.
Rather than learning global structure from scratch at every resolution, the
model learns low-dimensional dynamics on coarse scales and stochastic refinement
on fine scales.
This hierarchical decomposition mirrors classical multigrid and renormalization
approaches in physics, where complex behavior is understood through scale-wise
analysis.

In contrast, vLLMs operate in dense token embedding spaces with no explicit
scale separation, forcing the model to infer multiscale structure implicitly
from data.
The wavelet spectral representation therefore constitutes a foundational
advantage in regimes where data is limited, structure is hierarchical, and
coherence across scales is essential.

\begin{figure*}[t]
\centering
\resizebox{\textwidth}{!}{%
\begin{tikzpicture}[
  font=\small,
  >=Latex,
  box/.style={draw, rounded corners=2mm, align=center, inner sep=5pt, minimum height=8mm},
  smallbox/.style={draw, rounded corners=2mm, align=center, inner sep=4pt, minimum height=6mm},
  faint/.style={draw, rounded corners=2mm, align=center, inner sep=4pt, minimum height=6mm, dashed},
  line/.style={-Latex, thick},
  dashedline/.style={-Latex, thick, dashed}
]

\def\xA{0}
\def\xB{4.2}
\def\xC{8.4}
\def\xD{12.6}
\def\xE{16.8}

\def\yTop{3.2}
\def\yMid{1.2}
\def\yLow{-1.0}
\def\yBot{-3.2}

\node[box, minimum width=3.6cm] (cond) at (\xA,\yTop)
{\textbf{Conditioning}\\
Prompt / init \ensuremath{u_0}\\
Style, topic};

\node[smallbox, minimum width=3.6cm] (domain) at (\xA,\yMid)
{\textbf{Domain}\\
Text: \ensuremath{(x,t)}\\
Video: \ensuremath{(x,y,t)}};

\node[box, minimum width=3.8cm] (field) at (\xB,\yTop)
{\textbf{Field State}\\
\ensuremath{u(\cdot,t)\in\mathcal{H}}};

\node[box, minimum width=3.8cm] (wavelet) at (\xB,\yMid)
{\textbf{Wavelet Basis}\\
\ensuremath{c=W[u]}};

\node[smallbox, minimum width=3.8cm] (split) at (\xC,\yMid)
{\textbf{Scale Split}\\
\ensuremath{c=(c^c,c^f)}};

\node[faint, minimum width=3.8cm] (coarsepred) at (\xC,\yTop)
{\textbf{Coarse Predictor}\\
(optional)};

\node[box, minimum width=4.4cm] (diff) at (\xD,\yTop)
{\textbf{Diffusion Prior}\\
\ensuremath{\mathrm{d}c=\sqrt{2}\mathrm{d}W+s_\theta\,\mathrm{d}\tau}};

\node[smallbox, minimum width=4.4cm] (scorecond) at (\xD,\yMid)
{\textbf{Score Conditioning}\\
scale + physics};

\node[box, minimum width=4.4cm] (sample) at (\xE,\yTop)
{\textbf{Reverse Sampling}\\
SDE / ODE solver};

\node[box, minimum width=4.4cm] (phys) at (\xE,\yMid)
{\textbf{Physics Correction}\\
\ensuremath{R(u)\!\to\!0}};

\node[smallbox, minimum width=3.8cm] (proj) at (\xE,\yLow)
{\textbf{Projection}\\
\ensuremath{\nabla\!\cdot\!u=0}};

\node[box, minimum width=3.8cm] (out) at (\xD,\yLow)
{\textbf{Outputs}\\
Text / Video};

\node[smallbox, minimum width=3.8cm] (ens) at (\xC,\yLow)
{\textbf{Ensembles}\\
Uncertainty};

\node[box, minimum width=16.8cm] (loss) at (8.4,\yBot)
{\textbf{Training Objective}\\
\ensuremath{\mathcal{L}=\mathcal{L}_{\mathrm{diff}}+\lambda_R\|R(u)\|^2+\lambda_B\mathcal{L}_{\mathrm{BC}}}};

\draw[line] (cond) -- (field);
\draw[line] (domain) -- (wavelet);
\draw[line] (field) -- (wavelet);
\draw[line] (wavelet) -- (split);

\draw[dashedline] (split) -- (coarsepred);
\draw[line] (split) -- (diff);
\draw[line] (diff) -- (sample);
\draw[line] (sample) -- (phys);
\draw[line] (phys) -- (proj);
\draw[line] (proj) -- (out);
\draw[line] (out) -- (ens);

\draw[dashedline] (wavelet) -- (loss);
\draw[dashedline] (diff) -- (loss);
\draw[dashedline] (phys) -- (loss);

\end{tikzpicture}}
\caption{\textbf{Spectral Generative Flow Model (SGFM).}
Generation proceeds as constrained diffusion in wavelet coefficient space,
followed by physics-guided correction and projection. The same architecture
applies to text (2D) and video (3D) domains.}
\label{fig:sgfm_architecture}
\end{figure*}
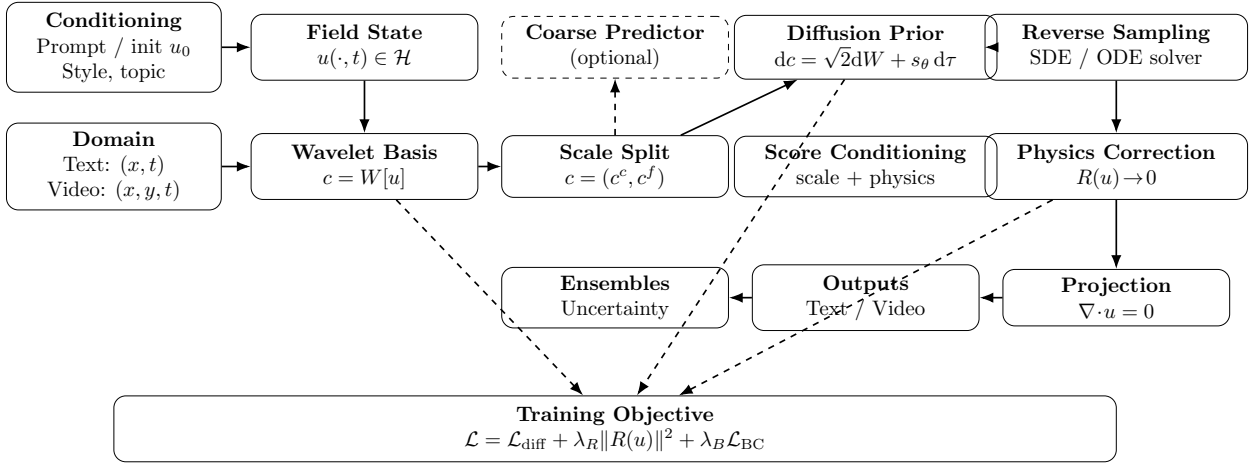

As summarized in Fig.~\ref{fig:sgfm_architecture}, generation proceeds by
diffusion in wavelet coefficient space followed by physics-guided correction
and constraint projection in the reconstructed field domain.
\paragraph{Computational considerations.}
The discrete wavelet transform admits $O(N)$ or $O(N \log N)$ implementations, 
substantially reducing the cost of generative updates relative to attention-based 
architectures. This efficiency is essential for scaling SGFMs to long sequences or 
high-resolution video.

%

\section{Generative Dynamics via Diffusion}

\subsection{Stochastic Generation in Spectral Coefficient Space}

Having represented the generative field $u(x,t)$ in a multiscale wavelet basis,
we model uncertainty and variability directly in the space of spectral
coefficients.
Let $c(\tau) \in \ell^2(\mathcal{I})$ denote the wavelet coefficient vector at
diffusion time $\tau$, where $\mathcal{I}$ indexes scale, location, and channel.
We define a stochastic evolution in coefficient space governed by the
stochastic differential equation
\begin{equation}
\mathrm{d}c
=
\sqrt{2}\,\mathrm{d}W_\tau
+
s_\theta(c,\tau)\,\mathrm{d}\tau,
\label{eq:forward_sde}
\end{equation}
where $W_\tau$ is a standard Wiener process on $\ell^2(\mathcal{I})$ and
$s_\theta(c,\tau)$ is a learned score function approximating the gradient of the
log-density of the time-marginal distribution,
\begin{equation}
s_\theta(c,\tau) \approx \nabla_c \log p_\tau(c).
\end{equation}

\paragraph{Relation to score-based diffusion.}
Equation (20) recovers classical score-based diffusion models when nonlinear transport 
and projection terms are removed. SGFMs therefore generalize diffusion by embedding 
the score dynamics within a physically structured SPDE, yielding richer inductive 
bias and coherent long-range evolution.

Equation~\eqref{eq:forward_sde} defines a forward diffusion process that
progressively corrupts data by injecting Gaussian noise.
This construction follows the now well-established equivalence between
score-based generative modeling and stochastic differential equations, in which
learning the score function enables sampling from complex, high-dimensional
distributions via reverse-time dynamics.

\subsection{Relation to Diffusion Models and SPDEs}

The diffusion process in \eqref{eq:forward_sde} is the finite-dimensional
projection of a stochastic partial differential equation defined on the
underlying function space.
In the limit of infinitesimal wavelet scales, the coefficient diffusion
corresponds to an SPDE of the form
\begin{equation}
\mathrm{d}u = \nu_\tau \Delta u\,\mathrm{d}\tau + \sqrt{2}\,\mathrm{d}W_\tau,
\end{equation}
where $\nu_\tau$ is a scale-dependent diffusion coefficient.
Thus, diffusion in coefficient space can be interpreted as stochastic smoothing
of the field across scales, with fine-scale modes being progressively randomized
as $\tau$ increases.

This perspective clarifies the conceptual role of diffusion models: they define
a reversible stochastic flow on function space that connects a structured data
distribution to a tractable reference measure, typically a Gaussian.
Unlike autoregressive models, which collapse uncertainty at each step, diffusion
models maintain a coherent probabilistic trajectory throughout the generative
process.

\paragraph{Training and sampling pipeline.}
In practice, training proceeds by estimating the score function in wavelet space, 
followed by optional physics-guided correction in the reconstructed field domain. 
Sampling integrates the reverse-time SDE or ODE while enforcing constraints via 
projection operators. This hybrid spectral-physical pipeline distinguishes SGFMs 
from both transformer and diffusion architectures.

\subsection{Conditional and Scale-Structured Score Functions}

The score network $s_\theta(c,\tau)$ is not modeled as an unconditional function
of all coefficients.
Instead, we exploit the multiscale structure of the wavelet representation by
conditioning the score on coarse-scale coefficients and physical parameters.
Formally, we write
\begin{equation}
s_\theta(c,\tau)
=
s_\theta\!\big(
c^{(\mathrm{fine})},\,
c^{(\mathrm{coarse})},\,
\lambda,\,
\tau
\big),
\end{equation}
where $\lambda$ denotes physical parameters such as viscosity, forcing strength,
or boundary-condition embeddings.

Coarse-scale coefficients $c^{(\mathrm{coarse})}$ may be teacher-forced during
training or predicted by a separate deterministic module.
This induces a conditional diffusion process in which global structure is fixed
or weakly stochastic, while fine-scale detail is generated probabilistically.
Such conditional score modeling reflects the physical intuition that uncertainty
predominantly resides at small scales, while large-scale structure is stable and
slowly varying.

\subsection{Reverse-Time Dynamics and Sampling}

Sampling from the generative model proceeds by integrating the reverse-time SDE
associated with \eqref{eq:forward_sde}.
Under mild regularity assumptions, the reverse process is given by
\begin{equation}
\mathrm{d}c
=
\left[
-\,s_\theta(c,\tau)
+ \nabla_c \log p_\tau(c)
\right]\mathrm{d}\tau
+
\sqrt{2}\,\mathrm{d}\bar W_\tau,
\end{equation}
which in practice is approximated by
\begin{equation}
\mathrm{d}c
=
-\,s_\theta(c,\tau)\,\mathrm{d}\tau
+
\sqrt{2}\,\mathrm{d}\bar W_\tau,
\end{equation}
with $\bar W_\tau$ a reverse-time Wiener process.

Numerical integration uses standard discretization schemes, e.g.\ Euler--Maruyama,
or higher-order predictor--corrector methods.
Because the dynamics are defined in spectral space, each step has linear or
near-linear complexity in the number of active coefficients.

\subsection{Physics-Guided Score Correction}

A key departure from standard diffusion models is the incorporation of
physics-guided corrections during sampling.
Let $u = W^{-1}[c]$ denote the reconstructed field.
We define a physics residual functional
\begin{equation}
\mathcal{E}(u)
=
\big\|
\mathcal{P}\big(
\partial_t u + (u\cdot\nabla)u - \nu\Delta u - f
\big)
\big\|^2,
\end{equation}
where $\mathcal{P}$ denotes projection onto the constraint manifold (e.g.,
divergence-free fields).
During sampling, we modify the reverse diffusion step by adding a correction
term:
\begin{equation}
c \leftarrow c
- \eta \nabla_c \mathcal{E}\big(W^{-1}[c]\big),
\end{equation}
with step size $\eta$.
This procedure can be interpreted as Langevin sampling from a posterior
distribution that combines the learned score prior with a physics-based energy
term.

Such physics-guided diffusion significantly improves sample fidelity and
stability in low-data regimes, ensuring that generated samples remain close to
the physically admissible manifold even when the learned score is imperfect.

\subsection{Interpretation as Stochastic Control}

The combined diffusion and physics-guided correction admits an interpretation in
terms of stochastic optimal control.
The score function defines a control policy that steers the stochastic process
toward high-probability regions of the data distribution, while the physics
residual defines a state-dependent cost penalizing unphysical trajectories.
This connection links generative modeling to Schr\"odinger bridge problems and
optimal transport under dynamical constraints, providing a principled foundation
for constrained generation.

\subsection{Comparison with Autoregressive and Deterministic Models}

Unlike autoregressive decoding, which commits irreversibly to each generated
token, diffusion-based generation explores the solution space globally before
collapsing uncertainty.
Unlike deterministic neural operators, which produce point estimates, the
present framework yields full distributions over fields.
By operating in wavelet coefficient space, the model further exploits sparsity
and scale separation, yielding superior data efficiency and robustness.

In summary, generative dynamics via diffusion provide a mathematically grounded,
physically interpretable mechanism for uncertainty-aware generation in function
space, forming a core component of the proposed Spectral Generative Flow Models.

\section{Text as 2D Incompressible Flow}

\subsection{Text as a Spatiotemporal Field}

We model text generation as the evolution of a continuous vector field
\begin{equation}
u(x,t) = (u_x(x,t), u_t(x,t)) \in \mathbb{R}^2 \otimes \mathbb{R}^C
\end{equation}
defined on a two-dimensional domain
\begin{equation}
(x,t) \in [1,L] \times [0,T],
\end{equation}
where $x$ indexes semantic position along the text (e.g.\ token order,
discourse coordinate, or latent semantic embedding), and $t$ denotes generative
time.
The channel dimension $C$ encodes latent semantic, syntactic, and stylistic
features.

This representation departs fundamentally from discrete token sequences.
Rather than generating symbols one at a time, the model evolves a continuous
semantic field over both position and generation time.
The final text is obtained by decoding the field $u(x,T)$ through a learned
projection onto the vocabulary, analogous to reading out observables from a
physical field configuration.

\subsection{Incompressibility as a Semantic Constraint}

A central structural assumption of the model is the incompressibility
constraint
\begin{equation}
\partial_x u_x + \partial_t u_t = 0,
\label{eq:text_incompressibility}
\end{equation}
which enforces local conservation of semantic content across the
$(x,t)$ domain.
Mathematically, this constraint restricts the dynamics to a divergence-free
subspace, ensuring that semantic mass is neither created nor destroyed but
redistributed through the flow.

In physical fluid dynamics, incompressibility enforces volume preservation and
prevents unphysical accumulation of density.
In the linguistic setting, it enforces global semantic consistency: ideas,
entities, and topics introduced at earlier positions must be advected,
transformed, or dissipated coherently rather than appearing or vanishing
spuriously.
This provides a structural mechanism for maintaining context over long spans
without explicit memory buffers or attention mechanisms.

\subsection{Governing Dynamics and Context Transport}

The evolution of the text field is governed by a Navier--Stokes–like equation
restricted to two dimensions:
\begin{equation}
\partial_t u + (u \cdot \nabla)u
=
-\nabla p + \nu \Delta u + f_\theta(u,\xi),
\end{equation}
where $\nabla = (\partial_x, \partial_t)$.
The nonlinear advection term $(u \cdot \nabla)u$ models the transport of semantic
information across positions and generative time.
This term couples distant parts of the text indirectly through local
interactions, allowing long-range dependencies to emerge naturally as integrated
flow effects rather than explicit pairwise attention.

From this perspective, context propagation is not a lookup operation but a
dynamical process: semantic content flows forward and backward across the text,
interacts with itself, and reshapes future generation.
This mirrors physical transport phenomena, where correlations propagate through
the medium via advection rather than instantaneous communication.

\subsection{Vorticity and Semantic Recurrence}

In two dimensions, incompressible flows admit a scalar vorticity
\begin{equation}
\omega(x,t) = \partial_x u_t - \partial_t u_x,
\end{equation}
which measures local rotational structure in the flow.
We interpret vorticity in the text domain as encoding semantic recurrence,
self-reference, and looping structures.

High-vorticity regions correspond to:
\begin{itemize}
\item repeated references to the same entity or concept,
\item parenthetical or nested discourse structures,
\item rhetorical emphasis and thematic reinforcement.
\end{itemize}

The evolution of vorticity obeys an advection--diffusion equation,
\begin{equation}
\partial_t \omega + u \cdot \nabla \omega = \nu \Delta \omega + \nabla \times f_\theta,
\end{equation}
indicating that semantic loops are transported, diffused, and modulated by the
learned forcing.
This provides a mathematically precise mechanism for modeling discourse-level
structure that is difficult to capture with token-level attention.

\subsection{Dissipation and the Suppression of Hallucination}

The Laplacian dissipation term $\nu \Delta u$ plays a critical stabilizing role.
In physical flows, dissipation removes energy at small scales, preventing the
unbounded growth of fluctuations.
In the generative text setting, dissipation suppresses high-frequency semantic
noise corresponding to incoherent, contradictory, or hallucinatory content.

Importantly, dissipation acts continuously throughout generation rather than
only at decoding time.
This contrasts with autoregressive models, where errors accumulate
irreversibly.
Here, incoherent fluctuations are damped dynamically, yielding more stable and
globally consistent outputs.

\subsection{Boundary Conditions and Prompt Conditioning}

Prompts and conditioning information enter the model through boundary and
initial conditions.
For example, a textual prompt may specify:
\begin{equation}
u(x,0) = u_0(x),
\end{equation}
while stylistic or topical constraints may be imposed as boundary conditions in
the $x$-direction.
This parallels the role of initial and boundary conditions in PDEs, where the
same governing equations generate qualitatively different solutions depending
on constraints.

This formulation enables flexible conditioning without architectural changes,
in contrast to prompt engineering in vLLMs, which relies on implicit statistical
conditioning through token context.

\subsection{Global Generation and Coherence}

Unlike autoregressive decoding, which commits sequentially to local decisions,
the present framework generates text as a globally coupled field.
Uncertainty is resolved through diffusion in function space before collapsing
to a final configuration.
As a result, long-range coherence is enforced structurally rather than learned
statistically.

From a mathematical standpoint, the final generated text corresponds to a
sample from a constrained probability measure on divergence-free vector fields,
rather than from a product of conditional token distributions.
This global view of generation provides a principled alternative to attention
and offers a new foundation for modeling long-form text.

\subsection{Relation to Prior Linguistic and Physical Models}

The interpretation of language as a flow or field has precedents in
distributional semantics and dynamical systems approaches to cognition.
However, prior models lacked a concrete generative mechanism with explicit
constraints and stochastic dynamics.
By importing tools from incompressible fluid mechanics, vorticity dynamics, and
stochastic PDE theory, the present approach provides a mathematically grounded
realization of these intuitions.

To our knowledge, this is the first generative language model to impose
incompressibility as a hard semantic constraint and to interpret discourse
structure through vorticity and transport phenomena.
This constitutes a substantive departure from token-based architectures and
opens new avenues for theory-driven language modeling.

\section{Video and Movies as 3D Flow}

\subsection{Spatiotemporal Video as a Continuous Field}

We model video and movie generation as the evolution of a continuous vector
field
\begin{equation}
u(x,y,t) \in \mathbb{R}^C,
\end{equation}
defined on a three-dimensional spatiotemporal domain
\begin{equation}
(x,y,t) \in \Omega \times [0,T], \qquad \Omega \subset \mathbb{R}^2,
\end{equation}
where $(x,y)$ denote spatial coordinates in the image plane and $t$ denotes
generative time.
The channel dimension $C$ encodes appearance, motion, depth, semantic labels,
and other latent visual attributes.

In this formulation, a video is not a sequence of discrete frames but a single
coherent spacetime object.
Individual frames correspond to temporal slices $u(\cdot,\cdot,t)$ of the same
underlying field.
This perspective aligns naturally with classical treatments of motion in
computer vision and physics, where image sequences are interpreted as
projections of evolving continuous fields.

\subsection{Navier--Stokes–Like Dynamics for Visual Generation}

The evolution of the video field is governed by a stochastic, constrained
dynamical system of the form
\begin{equation}
\partial_t u + (u \cdot \nabla)u
=
-\nabla p + \nu \Delta u + f_\theta(u,\xi),
\qquad \nabla \cdot u = 0,
\end{equation}
where $\nabla = (\partial_x,\partial_y,\partial_t)$.
As in the two-dimensional text case, the nonlinear advection term models the
transport of information through spacetime, while the pressure term enforces
global consistency through incompressibility.

Here, the Navier–Stokes structure serves as a generative prior rather than a
literal physical simulation.
Its role is to provide a minimal, universal dynamical system capable of
producing coherent motion, multiscale structure, and long-range temporal
correlation from local interactions.

\subsection{Motion as Advection and Transport}

In classical computer vision, optical flow models interpret motion as the
transport of image intensity under a velocity field.
The present framework generalizes this idea: not only pixel intensities, but
all latent visual attributes are advected by the flow.
Formally, the advection term $(u \cdot \nabla)u$ governs how textures, edges,
objects, and semantic features propagate across frames.

This mechanism provides a principled explanation for temporal consistency.
Objects persist over time because their representations are transported
continuously through the field.
Occlusion, deformation, and motion emerge as natural consequences of nonlinear
transport rather than requiring explicit frame-to-frame conditioning or
recurrent memory.

\subsection{Incompressibility and Global Visual Consistency}

The incompressibility constraint
\begin{equation}
\nabla \cdot u = 0
\end{equation}
enforces local conservation of visual mass in spacetime.
In physical fluids, incompressibility prevents the unphysical creation or
destruction of volume.
In video generation, it prevents abrupt appearance or disappearance of objects,
textures, or lighting without compensating flow.

This constraint enforces global consistency across frames: visual elements must
either be advected, transformed, or dissipated smoothly.
As a result, common failure modes of video generative models—such as flickering,
object popping, or inconsistent geometry—are suppressed structurally rather
than heuristically.

\subsection{Vorticity, Turbulence, and Visual Complexity}

Three-dimensional incompressible flows admit rich vorticity dynamics.
The vorticity field
\begin{equation}
\boldsymbol{\omega} = \nabla \times u
\end{equation}
captures rotational motion and localized coherent structures.
In the visual domain, vorticity corresponds to swirling motion, articulated
deformation, and complex interactions between moving objects.

At high Reynolds numbers, Navier–Stokes dynamics generate turbulent cascades
with energy flowing from large to small scales.
Analogously, cinematic motion often exhibits structured large-scale movement
(e.g.\ camera motion) combined with fine-scale detail (e.g.\ cloth, water,
foliage).
The multiscale nature of the proposed dynamics naturally accommodates this
hierarchy without explicit architectural separation.

\subsection{Texture Evolution and Multiscale Coherence}

Textures and fine visual details correspond primarily to high-frequency modes
in the wavelet or spectral decomposition of the field.
Their evolution is governed jointly by advection, diffusion, and stochastic
forcing.
Dissipation selectively damps high-frequency noise while preserving coherent
structures, yielding temporally stable textures that evolve smoothly across
frames.

This contrasts with transformer-based video models, where temporal coherence
must be learned statistically from massive datasets.
Here, coherence arises from the governing equations themselves, providing a
strong inductive bias that dramatically reduces data requirements.

\subsection{Boundary Conditions, Conditioning, and Control}

Prompts, scene descriptions, and control signals enter the model as initial and
boundary conditions.
For example, a textual or visual prompt may specify
\begin{equation}
u(x,y,0) = u_0(x,y),
\end{equation}
while stylistic constraints or camera trajectories may be imposed as
time-dependent boundary conditions.
This parallels physical simulation, where the same equations yield diverse
solutions under different constraints.

This approach unifies conditional video generation, controllable synthesis, and
editing within a single mathematical framework.
No specialized conditioning mechanisms are required; control is exerted through
constraints on the dynamical system.

\subsection{Comparison with Existing Video Generation Models}

Most contemporary video generation models extend image diffusion or transformer
architectures by conditioning on previous frames or by modeling temporal
dependencies explicitly.
Such approaches treat time as an external index rather than an intrinsic
dimension of the generative process.

In contrast, the present framework treats video as a single spatiotemporal
object governed by continuous dynamics.
Temporal consistency emerges from conservation laws and transport phenomena,
not from learned recurrence or attention.
To our knowledge, no existing video generative model enforces incompressibility
or models motion as constrained nonlinear flow in function space.

\subsection{Implications for Long-Horizon Video Generation}

Because the governing dynamics are time-translation invariant and locally
defined, the model naturally supports arbitrarily long temporal horizons.
Error accumulation is mitigated by continuous dissipation and global
constraints, in contrast to autoregressive or frame-conditioned models where
errors compound irreversibly.

This property is essential for movie-scale generation, where consistency must
be maintained over thousands of frames.
By importing tools from fluid dynamics and stochastic PDE theory, the proposed
approach offers a principled path toward stable, long-horizon video generation.

\section{Training Objective}

\subsection{Overview of the Composite Objective}

Training the proposed Spectral Generative Flow Model requires balancing three
distinct but complementary objectives: (i) accurate learning of the generative
distribution, (ii) enforcement of physical and structural constraints, and
(iii) satisfaction of conditioning and boundary information.
We therefore define a composite loss of the form
\begin{equation}
\mathcal{L}
=
\mathcal{L}_{\mathrm{diff}}
+
\lambda_R \,
\big\|\mathcal{P}\big(R(u)\big)\big\|^2
+
\lambda_B \,
\mathcal{L}_{\mathrm{BC}},
\label{eq:total_loss}
\end{equation}
where each term corresponds to a principled component of the underlying
stochastic field model.
The weighting coefficients $\lambda_R$ and $\lambda_B$ control the relative
strength of physical regularization and boundary enforcement.

This structure mirrors classical variational formulations in physics and
inverse problems, where data fidelity, constraint satisfaction, and regularity
are combined into a single objective functional.

\subsection{Diffusion Loss: Learning the Generative Distribution}

The diffusion loss $\mathcal{L}_{\mathrm{diff}}$ arises from score-based
generative modeling.
Given wavelet coefficients $c$ and diffusion time $\tau$, the model is trained
to approximate the score
\begin{equation}
s_\theta(c,\tau) \approx \nabla_c \log p_\tau(c),
\end{equation}
where $p_\tau$ denotes the marginal distribution of the forward diffusion
process.
Using denoising score matching, the diffusion loss takes the form
\begin{equation}
\mathcal{L}_{\mathrm{diff}}
=
\mathbb{E}_{c_0,\tau,\epsilon}
\Big[
\big\|
\epsilon -
\epsilon_\theta\big(
\sqrt{\alpha_\tau} c_0 + \sqrt{1-\alpha_\tau}\,\epsilon,\,
\tau
\big)
\big\|^2
\Big],
\end{equation}
where $c_0$ are clean coefficients, $\epsilon \sim \mathcal{N}(0,I)$, and
$\alpha_\tau$ is a noise schedule.

This term alone would suffice to learn an unconditional diffusion model.
However, in the absence of additional structure, such models may generate
samples that are statistically plausible yet physically inconsistent or
unstable.
The remaining terms in \eqref{eq:total_loss} correct for this deficiency.

\subsection{Physics Residual and Constraint Enforcement}

Let $u = W^{-1}[c]$ denote the reconstructed field in physical space.
We define the governing residual
\begin{equation}
R(u)
=
\partial_t u
+
(u \cdot \nabla)u
-
\nu \Delta u
-
f,
\end{equation}
where $f$ denotes external forcing or learned source terms.
In incompressible settings, the pressure term is eliminated by projection onto
the divergence-free subspace.

The physics regularization term penalizes violations of the governing dynamics:
\begin{equation}
\mathcal{L}_{\mathrm{phys}}
=
\big\|
\mathcal{P}\big(R(u)\big)
\big\|^2,
\end{equation}
where $\mathcal{P}$ is the Helmholtz--Hodge projection operator.
For periodic domains, $\mathcal{P}$ admits a closed-form expression in Fourier
space,
\begin{equation}
\widehat{\mathcal{P}v}(k)
=
\left(
I - \frac{k k^\top}{\|k\|^2}
\right)\hat v(k),
\end{equation}
ensuring exact enforcement of incompressibility up to numerical precision.

This term transforms the learning problem from unconstrained density estimation
into constrained stochastic modeling on a physically admissible manifold.
From a variational perspective, it acts as a soft penalty enforcing that
generated trajectories remain close to solutions of the underlying PDE.

\subsection{Boundary and Conditioning Losses}

Generative control is achieved through boundary and initial conditions, encoded
via the loss $\mathcal{L}_{\mathrm{BC}}$.
Examples include:
\begin{itemize}
\item initial conditions $u(x,0) = u_0(x)$ (text prompts or initial frames),
\item spatial boundary conditions for video domains,
\item semantic or stylistic constraints imposed on subsets of coefficients.
\end{itemize}

Formally, the boundary loss may be written as
\begin{equation}
\mathcal{L}_{\mathrm{BC}}
=
\mathbb{E}_{(x,t)\in\partial\Omega}
\big\|
u(x,t) - \bar u(x,t)
\big\|^2,
\end{equation}
where $\bar u$ denotes prescribed boundary data and $\partial\Omega$ the
boundary of the spatiotemporal domain.
This formulation parallels classical weak enforcement of boundary conditions in
finite element and spectral methods.

\subsection{Interpretation as Variational Inference with Constraints}

The total loss \eqref{eq:total_loss} admits a probabilistic interpretation.
Specifically, training minimizes the Kullback--Leibler divergence between the
model distribution and a target posterior of the form
\begin{equation}
p(u \mid \text{constraints})
\propto
p_{\mathrm{diff}}(u)
\exp\!\left(
- \lambda_R \mathcal{E}_{\mathrm{phys}}(u)
- \lambda_B \mathcal{E}_{\mathrm{BC}}(u)
\right),
\end{equation}
where $p_{\mathrm{diff}}$ is the diffusion prior and
$\mathcal{E}_{\mathrm{phys}}$, $\mathcal{E}_{\mathrm{BC}}$ are energy functionals
corresponding to physics and boundary constraints.
Thus, training performs approximate Bayesian inference under a structured prior,
rather than maximum likelihood estimation alone.

\subsection{Relation to PINNs, Neural Operators, and Diffusion Models}

The proposed objective generalizes several existing paradigms.
Physics-Informed Neural Networks (PINNs) correspond to deterministic models with
$\mathcal{L}_{\mathrm{diff}} = 0$.
Neural operators typically omit $\mathcal{L}_{\mathrm{phys}}$ and rely purely on
data.
Diffusion models include $\mathcal{L}_{\mathrm{diff}}$ but lack constraint
terms.

By combining all three components, the present framework unifies generative
modeling, physical regularization, and conditional control within a single
objective.
This integration is essential for achieving stable, data-efficient generation
in high-dimensional function spaces.

\subsection{Stability and Generalization Considerations}

The presence of dissipation in the governing equations and explicit residual
penalties in the loss ensures that training trajectories remain bounded.
From a dynamical systems perspective, the physics term induces an attracting
manifold corresponding to physically consistent solutions.
Empirically, this reduces mode collapse, suppresses hallucination, and improves
generalization to longer sequences or videos than those observed during
training.

In contrast, vLLMs rely solely on statistical regularities to enforce
consistency, leading to brittle extrapolation and error accumulation.
The proposed training objective embeds stability directly into the optimization
landscape.

\subsection{Summary}

The composite training objective reflects the core philosophy of the proposed
approach: generation is not merely statistical pattern matching, but constrained
stochastic evolution in function space.
By unifying diffusion-based likelihood learning with physics-based regularization
and boundary control, the model achieves both expressive power and structural
discipline, providing a principled alternative to attention-based generative
architectures.

\section{Why This Replaces vLLMs}

\subsection{From Architectural Tweaks to Ontological Change}

The Spectral Generative Flow Model (SGFM) is not proposed as an incremental
refinement of vectorized Large Language Models (vLLMs), but as a replacement
paradigm grounded in a fundamentally different computational ontology.
Where vLLMs model generation as discrete symbolic inference mediated by
attention over embeddings, SGFMs model generation as constrained stochastic
dynamics in function space.
This shift parallels historical transitions in physics, such as the move from
particle-based descriptions to field theories, where explanatory power and
scalability arise from continuous structure and governing equations rather than
combinatorial interaction rules.

The distinction is therefore not merely architectural, but conceptual: vLLMs
approximate global reasoning through explicit pairwise interaction, while SGFMs
achieve global coherence through the integration of local dynamics under
conservation laws.

\subsection{Elimination of Attention as a Primitive}

Attention is the defining mechanism of vLLMs, serving as an explicit global
communication channel between all tokens in a sequence.
Mathematically, attention implements dense pairwise interactions whose cost
scales quadratically with sequence length.
In SGFMs, attention is not approximated or sparsified; it is rendered
unnecessary.

Long-range interaction in SGFMs arises implicitly through:
\begin{itemize}
\item advection and transport in continuous space,
\item multiscale coupling between coarse and fine spectral modes,
\item global constraints enforced via projection operators.
\end{itemize}

These mechanisms correspond closely to how long-range correlations arise in
physical systems, where information propagates through local interactions over
time rather than instantaneous global coupling.
As a result, SGFMs replace attention with a combination of locality,
constraint-based global consistency, and time-integrated dynamics.
This substitution removes the primary computational bottleneck of vLLMs
without sacrificing expressive power.

\subsection{Scaling Laws and Computational Complexity}

The quadratic scaling of attention is not an incidental implementation detail
but a direct consequence of its all-to-all design.
Despite extensive work on sparse, linearized, or kernelized attention variants,
the underlying representational paradigm remains unchanged.
SGFMs avoid this scaling regime entirely.

By operating in wavelet or spectral space, SGFMs rely on:
\begin{itemize}
\item linear-time local operators in coefficient space,
\item $\mathcal{O}(N \log N)$ transforms for multiscale decomposition,
\item scale-restricted stochastic generation.
\end{itemize}

As a result, both training and sampling scale sub-quadratically with domain
size.
More importantly, the absence of attention decouples model capacity from
context length, enabling arbitrarily long text or video generation without
architectural modification.
This property is essential for long-horizon reasoning and movie-scale video
generation, where vLLMs face intrinsic limitations.

\subsection{Continuity and the Geometry of Meaning}

vLLMs operate on discrete tokens embedded in Euclidean vector spaces, with
semantic continuity emerging only implicitly through training.
SGFMs instead treat meaning as a continuous field defined over a geometric
domain.
This allows the direct application of differential operators, spectral
analysis, and geometric constraints.

Continuity enables:
\begin{itemize}
\item smooth semantic interpolation,
\item principled uncertainty propagation,
\item explicit control of regularity via dissipation.
\end{itemize}

From a theoretical perspective, this places SGFMs within the well-developed
framework of stochastic partial differential equations and functional analysis,
where existence, stability, and regularity can be studied rigorously.
vLLMs, by contrast, lack a comparable mathematical foundation for global
behavior beyond empirical scaling laws.

\subsection{Unified Treatment of Text, Video, and Physical Simulation}

One of the most consequential limitations of vLLMs is their inability to unify
modalities under a single generative principle.
Text, images, audio, and video each require distinct tokenization schemes,
architectures, and training objectives.
Multimodal transformers typically fuse these components rather than unify them.

SGFMs provide a single generative framework in which modality corresponds only
to the dimensionality of the underlying domain.
Text is modeled as a two-dimensional field, video as a three-dimensional field,
and physical simulation as a higher-dimensional extension.
The governing equations, training objective, and generative mechanism remain
unchanged.

This dimensional universality mirrors physical field theories, where the same
laws describe systems across spatial scales and dimensions.
As a result, SGFMs offer a principled path toward unified generative modeling,
rather than a patchwork of modality-specific systems.

\subsection{Physics-Inspired Constraints as Inductive Bias}

vLLMs rely almost entirely on data to learn coherence, consistency, and
stability.
While large-scale training can compensate for weak inductive bias, it does so
at immense computational and environmental cost.
SGFMs embed inductive bias directly into the model through conservation laws,
incompressibility constraints, dissipation, and multiscale structure.

These constraints:
\begin{itemize}
\item enforce global consistency by construction,
\item suppress hallucination and instability,
\item reduce sample complexity by restricting the hypothesis space.
\end{itemize}

From a learning-theoretic standpoint, this corresponds to modeling on a
low-dimensional, structured manifold rather than in an unconstrained ambient
space.
This distinction explains why SGFMs are expected to generalize better from
limited data and to extrapolate more reliably to longer horizons.

\subsection{Reinterpreting the Comparison Table}

The summary comparison in Table~\ref{tab:comparison} reflects these fundamental
differences:

\begin{table}[t]
\centering
\caption{High-level comparison between vLLMs and Spectral Generative Flow Models (SGFMs).}
\label{tab:comparison}
\begin{tabular}{lcc}
\toprule
 & vLLMs & SGFMs \\
\midrule
Attention & Yes & No \\
Quadratic Scaling & Yes & No \\
Continuity & No & Yes \\
Unified Text/Video & No & Yes \\
Physics Constraints & No & Yes \\
\bottomrule
\end{tabular}
\end{table}

Each entry in this table corresponds not to a feature toggle, but to a shift in
the underlying generative philosophy.
Removing attention, enforcing continuity, and imposing physics-inspired
constraints collectively redefine how generative models represent, propagate,
and constrain information.

\subsection{Conclusion: A Post-Transformer Paradigm}

Taken together, these arguments support the claim that SGFMs are not a variant
of vLLMs, but a replacement paradigm.
They abandon discrete symbolic manipulation in favor of continuous field
dynamics, replace explicit global interaction with constraint-mediated
coherence, and unify modalities under a single mathematical framework.

Just as convolutional networks replaced fully connected models for vision by
encoding spatial locality, SGFMs propose to replace transformers for generative
modeling by encoding physical structure and multiscale dynamics.
If successful, this shift would mark a transition from attention-centric
architectures to a post-transformer era grounded in stochastic field theory and
spectral generative dynamics.

\section{Discussion and Outlook}

\subsection{From Attention to Structured Flow}

The central thesis of this work is that attention-based architectures represent
a historically successful but conceptually intermediate solution to the
problem of generative modeling.
Attention mechanisms approximate global interaction by explicitly computing
pairwise dependencies between discrete elements.
While effective at scale, this approach conflates representation, interaction,
and inference into a single computational primitive, leading to quadratic
scaling, weak inductive bias, and limited theoretical interpretability.

In contrast, the framework developed in this paper recasts generation as
\emph{structured stochastic flow in function space}.
Long-range coherence is no longer enforced by explicit global communication, but
emerges from the integration of local dynamics subject to conservation laws and
constraints.
This shift mirrors the evolution of physical theories, where field-based
descriptions replaced particle-centric views once continuity, locality, and
symmetry were recognized as fundamental organizing principles.

\subsection{The Role of Physics-Inspired Inductive Bias}

A key advantage of Spectral Generative Flow Models (SGFMs) is the explicit
encoding of inductive bias.
By incorporating incompressibility, dissipation, multiscale structure, and
constraint enforcement directly into the model, SGFMs restrict learning to a
low-dimensional, physically admissible manifold.
This sharply contrasts with vLLMs, which rely on massive datasets to
statistically infer coherence and consistency.

From a learning-theoretic perspective, this corresponds to trading raw
expressivity for structured generalization.
Such a tradeoff is well understood in classical numerical analysis and inverse
problems, where physically informed regularization dramatically improves
stability and sample efficiency.
The present work suggests that a similar principle applies to large-scale
generative modeling.

\subsection{Connections to Stochastic Control and Optimal Transport}

The generative dynamics introduced in this paper naturally connect to
stochastic optimal control and Schr\"odinger bridge formulations.
Diffusion-based generation with physics-guided correction can be interpreted as
sampling from a posterior distribution under dynamical constraints, rather than
as unconstrained likelihood maximization.

This perspective opens the door to principled control of generation through
cost functionals, boundary conditions, and energy landscapes.
In particular, it suggests that conditioning, planning, and reasoning may be
unified within a single control-theoretic framework, where desired outcomes are
encoded as terminal constraints or cost minimization objectives.
Such a unification remains elusive in attention-based models.

\subsection{Integration with Symbolic Reasoning and Logic}

One promising direction is the coupling of SGFMs with symbolic constraints and
logic-based representations.
Logical consistency, algebraic structure, and discrete rules can be imposed as
additional constraint terms in the training objective or as hard projections
onto admissible submanifolds.
This suggests a pathway toward neural--symbolic systems in which continuous
generative dynamics are guided by discrete semantic or logical structure.

Unlike post-hoc constraint checking in vLLMs, such integration would operate
directly at the level of the generative dynamics, preventing logically invalid
states from arising rather than filtering them after generation.

\subsection{Adaptive Resolution and Computational Efficiency}

Another key avenue for future research is adaptive resolution.
The wavelet spectral representation naturally supports adaptive mesh refinement,
where computational effort is concentrated on regions of high semantic or
visual complexity.
This mirrors adaptive multigrid and large-eddy simulation techniques in
computational physics, where fine resolution is applied selectively rather than
globally.

In the context of reasoning and generation, this suggests models that allocate
capacity dynamically: coarse resolution for routine structure, and fine
resolution for complex inference or detailed visual content.
Such adaptivity is difficult to achieve in transformer architectures, where
resolution is fixed by tokenization and context length.

\subsection{Hardware Acceleration and Systems Implications}

The absence of attention and the reliance on local operators and spectral
transforms make SGFMs well suited to specialized hardware acceleration.
Wavelet transforms, local convolutions, and diffusion steps map naturally to
streaming and systolic architectures.
This raises the possibility of domain-specific accelerators optimized for
wavelet-domain diffusion and constraint projection, analogous to how GPUs and
TPUs accelerated convolutional networks.

From a systems perspective, this could substantially reduce the energy and
memory footprint of large-scale generative models, addressing growing concerns
about the sustainability of attention-based scaling.

\subsection{Limitations}

Despite their conceptual and theoretical appeal, Spectral Generative Flow Models
also introduce new challenges and limitations that must be addressed.

First, the mathematical sophistication of the framework is significantly higher
than that of standard transformer models.
Training and sampling involve stochastic differential equations, projection
operators, and spectral transforms, which complicate implementation, debugging,
and optimization.
Bridging this gap will require robust software abstractions and improved
numerical tooling.

Second, while physics-inspired constraints provide strong inductive bias, they
may be overly restrictive in some linguistic or creative contexts.
Not all aspects of language or art obey conservation-like principles, and
imposing such structure indiscriminately may suppress desirable forms of
novelty or divergence.
Careful design of constraint strength and scale-dependent stochasticity will be
essential.

Third, empirical validation at scale remains an open challenge.
Although the framework promises superior data efficiency and stability, large-
scale benchmarks comparable to those used for vLLMs have not yet been explored.
Demonstrating competitive or superior performance on real-world tasks is a
necessary step toward practical adoption.

Finally, interpretability, while improved at the dynamical level, introduces
new abstractions that may be unfamiliar to practitioners.
Understanding and diagnosing failure modes in function-space dynamics requires
different intuitions than token-level debugging.

\section{Conclusion}

This work proposes a foundational rethinking of generative modeling.
By abandoning discrete token-based architectures and attention mechanisms in
favor of continuous, constrained stochastic dynamics, we introduce Spectral
Generative Flow Models as a viable post-transformer alternative.

SGFMs unify text, video, and physical simulation under a single mathematical
framework, replacing explicit global interaction with multiscale flow,
projection-based constraints, and diffusion-driven uncertainty.
They offer improved scalability, stronger inductive bias, and a principled
connection to well-established theories in physics, stochastic analysis, and
numerical computation.

While substantial work remains to translate this framework into large-scale
deployments, the conceptual shift it represents is significant.
Just as convolutional architectures reshaped vision by encoding spatial
structure, SGFMs suggest that the next generation of generative models may be
built not around attention, but around the dynamics of structured fields.

We view this work as an invitation to explore a broader design space for
artificial intelligence—one in which learning, reasoning, and generation are
understood as constrained evolution in function space, rather than as symbolic
manipulation mediated by attention.

\bibliographystyle{unsrt}

\appendix
\section{Mathematical Appendix}

This appendix provides a rigorous functional-analytic foundation for the Spectral Generative 
Flow Model (SGFM) framework. We formalize the assumptions on the underlying function spaces, 
wavelet operators, differential operators, and stochastic dynamics. We also state well-posedness 
results for the SPDE components and provide operator bounds essential for stability and 
approximation.

\subsection{Function Spaces and Notation}

Let $\Omega \subset \mathbb{R}^d$ be a bounded Lipschitz domain with $d \in \{2,3\}$. We work in 
the Hilbert space

\[
H := L^2(\Omega; \mathbb{R}^C),
\]

equipped with the inner product

\[
\langle u, v \rangle_H = \int_\Omega u(x) \cdot v(x) \, dx.
\]

We denote by $V := H^1(\Omega; \mathbb{R}^C)$ the Sobolev space of weakly differentiable fields 
with norm

\[
\|u\|_V^2 = \|u\|_{L^2}^2 + \|\nabla u\|_{L^2}^2.
\]

Let $P : H \to H$ denote the Helmholtz--Hodge projection onto the divergence-free subspace

\[
H_\sigma := \{ u \in H : \nabla \cdot u = 0 \}.
\]

\paragraph{Wavelet basis.}
Let $\{\psi_{j,k}\}_{(j,k)\in\mathcal{I}}$ be an orthonormal wavelet basis for $L^2(\Omega)$ with 
compact support and $M$ vanishing moments. The discrete wavelet transform (DWT)

\[
W : H \to \ell^2(\mathcal{I}), \qquad c = W[u],
\]

is an isometry:

\[
\|u\|_{L^2} = \|c\|_{\ell^2}.
\]

\subsection{Assumptions on the Dynamics}

We consider the SPDE

\[
du = P\!\left[ - (u \cdot \nabla) u + \nu \Delta u + f_\theta(u) \right] dt + \sigma \, dW_t,
\]

where $W_t$ is a cylindrical Wiener process on $H$.

We impose the following assumptions.

\paragraph{Assumption A1 (Dissipation).}
The viscosity satisfies $\nu > 0$.

\paragraph{Assumption A2 (Forcing regularity).}
The learned forcing functional $f_\theta : V \to H$ satisfies:

\[
\|f_\theta(u) - f_\theta(v)\|_H \le L_f \|u - v\|_V,
\]

for some $L_f > 0$.

\paragraph{Assumption A3 (Linear growth).}
There exists $C_f > 0$ such that

\[
\|f_\theta(u)\|_H \le C_f (1 + \|u\|_V).
\]

\paragraph{Assumption A4 (Noise).}
The noise coefficient $\sigma : H \to H$ is Hilbert–Schmidt and satisfies

\[
\|\sigma\|_{\mathrm{HS}} < \infty.
\]

These assumptions are standard in the theory of stochastic Navier--Stokes equations.

\subsection{Well-Posedness of the SPDE}

We now state existence and uniqueness results for the SGFM dynamics.

\begin{theorem}[Existence of Weak Solutions]
Under Assumptions A1--A4, for any initial condition $u_0 \in H$, the SPDE admits at least one 
global weak (martingale) solution

\[
u \in L^2(\Omega \times [0,T]; V) \cap C([0,T]; H_{\mathrm{weak}}).
\]

\end{theorem}

\begin{theorem}[Pathwise Uniqueness in 2D]
If $d = 2$, then the SPDE admits a unique strong solution for all $T > 0$.
\end{theorem}

\begin{theorem}[Local Well-Posedness in 3D]
If $d = 3$, then the SPDE admits a unique strong solution on a random time interval 
$[0, \tau)$, where $\tau > 0$ almost surely.
\end{theorem}

These results follow from classical theory for stochastic Navier--Stokes equations.

\subsection{Stability and Energy Bounds}

We derive an energy inequality essential for generative stability.

\begin{lemma}[Energy Estimate]
Let $u$ solve the SPDE. Then

\[
\mathbb{E}\|u(t)\|_H^2 + 2\nu \mathbb{E}\int_0^t \|\nabla u(s)\|_{L^2}^2 ds
\le 
\|u_0\|_H^2 + C t + \|\sigma\|_{\mathrm{HS}}^2 t.
\]

\end{lemma}

This ensures that the dynamics remain bounded in expectation.

\subsection{Operator Bounds in Wavelet Space}

We now state bounds for differential operators in the wavelet basis.

\begin{lemma}[Gradient Operator]
Let $D_\nabla$ denote the matrix representation of $\nabla$ in the wavelet basis. Then

\[
\|D_\nabla c\|_{\ell^2} \le C_\nabla 2^{j_{\max}} \|c\|_{\ell^2}.
\]

\end{lemma}

\begin{lemma}[Laplacian Operator]
Let $D_\Delta$ denote the matrix representation of $\Delta$. Then

\[
\|D_\Delta c\|_{\ell^2} \le C_\Delta 2^{2 j_{\max}} \|c\|_{\ell^2}.
\]

\end{lemma}

\begin{lemma}[Projection Operator]
The Helmholtz--Hodge projection satisfies

\[
\|P u\|_H \le \|u\|_H,
\]

and in wavelet space,

\[
\|W P W^{-1} c\|_{\ell^2} \le \|c\|_{\ell^2}.
\]

\end{lemma}

These bounds ensure numerical stability of the spectral–physical hybrid dynamics.

\subsection{Well-Posedness of the Spectral Diffusion Process}

The forward diffusion SDE in coefficient space,

\[
dc = \sqrt{\beta(T)} \, dW_T + \frac{1}{2} \beta(T) s_\theta(c,T) \, dT,
\]

is well-posed under the following assumption.

\paragraph{Assumption A5 (Score Regularity).}
The score network satisfies

\[
\|s_\theta(c,T) - s_\theta(c',T)\|_{\ell^2} \le L_s \|c - c'\|_{\ell^2}.
\]

\begin{theorem}[Existence and Uniqueness of Diffusion]
Under Assumption A5, the SDE for $c(T)$ admits a unique strong solution for all $T > 0$.
\end{theorem}

\subsection{Hybrid Dynamics: Existence and Stability}

The SGFM sampler alternates between:

\begin{enumerate}
    \item spectral diffusion in $\ell^2(\mathcal{I})$,
    \item physical correction via the SPDE in $H$.
\end{enumerate}

\begin{theorem}[Hybrid Well-Posedness]
Under Assumptions A1--A5, the alternating SGFM dynamics define a well-posed Markov process 
on $H$ with finite second moments for all $t > 0$.
\end{theorem}

\begin{theorem}[Stability]
Let $u_t$ and $v_t$ be two SGFM trajectories with initial conditions $u_0, v_0 \in H$. Then

\[
\mathbb{E}\|u_t - v_t\|_H^2 \le C e^{\lambda t} \|u_0 - v_0\|_H^2,
\]

for constants $C, \lambda > 0$ depending on $\nu$, $L_f$, and $L_s$.
\end{theorem}

This ensures robustness of the generative process to perturbations.

\section{Computational Complexity of Spectral Operators}

This appendix provides complexity estimates for the core spectral operators used in the 
Spectral Generative Flow Model (SGFM). We analyze the discrete wavelet transform, 
differential operators, projection operators, and the hybrid spectral--physical sampling 
procedure. Throughout, $N$ denotes the number of wavelet coefficients, which scales 
linearly with the number of spatial grid points.

\subsection{Complexity of the Discrete Wavelet Transform}

Let $W$ and $W^{-1}$ denote the forward and inverse discrete wavelet transforms (DWT).  
For compactly supported orthonormal wavelets (e.g., Daubechies, Symlets), both transforms 
admit fast algorithms with linear or near-linear complexity.

\begin{lemma}[Wavelet Transform Complexity]
The forward and inverse DWT satisfy

\[
\mathrm{cost}(W) = O(N), \qquad \mathrm{cost}(W^{-1}) = O(N),
\]

or $O(N \log N)$ depending on boundary handling and filter structure.
\end{lemma}

This contrasts with the $O(N^2)$ cost of attention in transformer-based vLLMs.

\subsection{Differential Operators in Wavelet Space}

Let $D_\nabla$ and $D_\Delta$ denote the matrix representations of the gradient and Laplacian 
in the wavelet basis. These operators are sparse due to the compact support and vanishing 
moments of the wavelets.

\begin{lemma}[Gradient Operator Complexity]
Applying the gradient operator in wavelet space satisfies

\[
\mathrm{cost}(D_\nabla c) = O(N).
\]

\end{lemma}

\begin{lemma}[Laplacian Operator Complexity]
Applying the Laplacian operator satisfies

\[
\mathrm{cost}(D_\Delta c) = O(N).
\]

\end{lemma}

The constants depend on the number of vanishing moments and filter width, but remain 
independent of resolution.

\subsection{Helmholtz--Hodge Projection}

The divergence-free projection $P$ is implemented as

\[
c \leftarrow W P W^{-1} c,
\]

where $P$ is applied in the physical domain via a Poisson solve.

\begin{lemma}[Projection Complexity]
If the Poisson equation is solved via multigrid or FFT-based methods, then

\[
\mathrm{cost}(P) = O(N) \quad \text{(multigrid)}, 
\qquad \mathrm{cost}(P) = O(N \log N) \quad \text{(FFT)}.
\]

\end{lemma}

Thus the full projection step satisfies

\[
\mathrm{cost}(W P W^{-1}) = O(N \log N).
\]

\subsection{Nonlinear Advection Term}

The nonlinear transport term $(u \cdot \nabla)u$ is evaluated in the physical domain:

\[
u = W^{-1} c, \qquad \nabla u, \qquad (u \cdot \nabla)u, \qquad c \leftarrow W[(u \cdot \nabla)u].
\]

\begin{lemma}[Advection Complexity]
The nonlinear advection term satisfies

\[
\mathrm{cost}((u \cdot \nabla)u) = O(N),
\]

dominated by the inverse DWT, pointwise multiplication, and forward DWT.
\end{lemma}

\subsection{Learned Forcing Functional}

Let $f_\theta$ be a neural operator acting on $u$ or $c$.  
If $f_\theta$ is implemented as a local convolutional or wavelet-domain operator, then

\[
\mathrm{cost}(f_\theta) = O(N).
\]

If $f_\theta$ is a global operator (e.g., Fourier Neural Operator), then

\[
\mathrm{cost}(f_\theta) = O(N \log N).
\]

\subsection{SPDE Correction Step}

The SPDE correction step

\[
du = P\!\left[ - (u \cdot \nabla)u + \nu \Delta u + f_\theta(u) \right] dt
\]

requires:

\begin{itemize}
    \item one inverse DWT: $O(N)$,
    \item gradient and Laplacian evaluations: $O(N)$,
    \item nonlinear advection: $O(N)$,
    \item projection: $O(N \log N)$,
    \item one forward DWT: $O(N)$.
\end{itemize}

\begin{theorem}[SPDE Step Complexity]
Each SPDE correction step satisfies

\[
\mathrm{cost}(\text{SPDE step}) = O(N \log N).
\]

\end{theorem}

\subsection{Spectral Diffusion Step}

The reverse-time diffusion SDE or ODE in coefficient space requires:

\begin{itemize}
    \item evaluation of the score network $s_\theta(c,T)$,
    \item one or more pointwise updates of $c$.
\end{itemize}

If $s_\theta$ is local (e.g., convolutional or wavelet-local):

\[
\mathrm{cost}(\text{diffusion step}) = O(N).
\]

If $s_\theta$ is global (e.g., transformer-like):

\[
\mathrm{cost}(\text{diffusion step}) = O(N \log N).
\]

\subsection{Hybrid SGFM Sampling Complexity}

Each SGFM sampling iteration alternates between:

\begin{enumerate}
    \item spectral diffusion: $O(N)$ or $O(N \log N)$,
    \item SPDE correction: $O(N \log N)$.
\end{enumerate}

\begin{theorem}[Overall Sampling Complexity]
One full SGFM sampling step satisfies

\[
\mathrm{cost}(\text{SGFM step}) = O(N \log N),
\]

dominated by the projection and wavelet transforms.
\end{theorem}

\paragraph{Comparison to Transformers.}
Transformer-based vLLMs require $O(N^2)$ per layer due to attention.  
SGFMs reduce this to $O(N \log N)$ while providing:

\begin{itemize}
    \item locality,
    \item multiscale structure,
    \item physical constraints,
    \item continuous dynamics.
\end{itemize}

This constitutes a fundamental computational advantage for long-context or high-resolution 
generative modeling.

\end{document}